\documentclass{article}
\usepackage{spconf,amsmath,graphicx}
\usepackage{float}
\usepackage{CJK}
\usepackage{color}
\usepackage{soul}

\title{Hybrid coarse-fine classification for head pose estimation}

\name{Haofan Wang$^{1,2}$, Zhenghua Chen$^{2}$, Yi Zhou$^{1}$}
\address{School of Information Science and Technology, Dalian Maritime University$^{1}$, Horizon Robotics$^{2}$}

\begin{document}
\begin{CJK*}{UTF8}{gkai}
%
\maketitle

\begin{abstract}
Head pose estimation, which computes the intrinsic Euler angles (yaw, pitch, roll) from the human, is crucial for gaze estimation, face alignment and 3D reconstruction. Traditional approaches heavily relies on the accuracy of facial landmarks. It limits their performances, especially when the visibility of face is not in good conditions. In this paper, to do the estimation without facial landmarks,  we combine the coarse and fine regression output together for a deep network.  Utilizing more quantization units for the angles, a fine classifier is trained with the help of other auxiliary coarse units. Integrating regression is adopted to get the final prediction. The proposed approach is evaluated on three challenging benchmarks. It achieves the state-of-the-art on AFLW2000, BIWI and performs favorably on AFLW. Code has been released on Github.\renewcommand{\thefootnote}{\fnsymbol{footnote}}\footnote{https://github.com/haofanwang/accurate-head-pose\label{web}}

\end{abstract}


\section{Introduction}
\label{sec:intro}
    Facial expression recognition is one of the most successful applications of convolutional neural network in the past few years. Recently, more and more attentions have been posed on 3D facial understanding. Most of existed methods of 3D understanding require extracting 2D facial landmarks. Establishing the corresponding relationship between 2D landmarks and a standardized 3D head model, 3D pose estimation of the head can be viewed as a by-product of the 3D understanding. While facial landmark detection has been improved by large scale, thanks to the deep neural network, the two-step based head pose estimation may have two extra errors.
    For example, poor facial landmark detection in the bad visual conditions harms the precision of angle estimation. Also, the precision of ad-hoc fitting of the 3D head model affects the accuracy of the pose estimation.

\begin {figure}[H]
\centering
\includegraphics[scale=0.09]{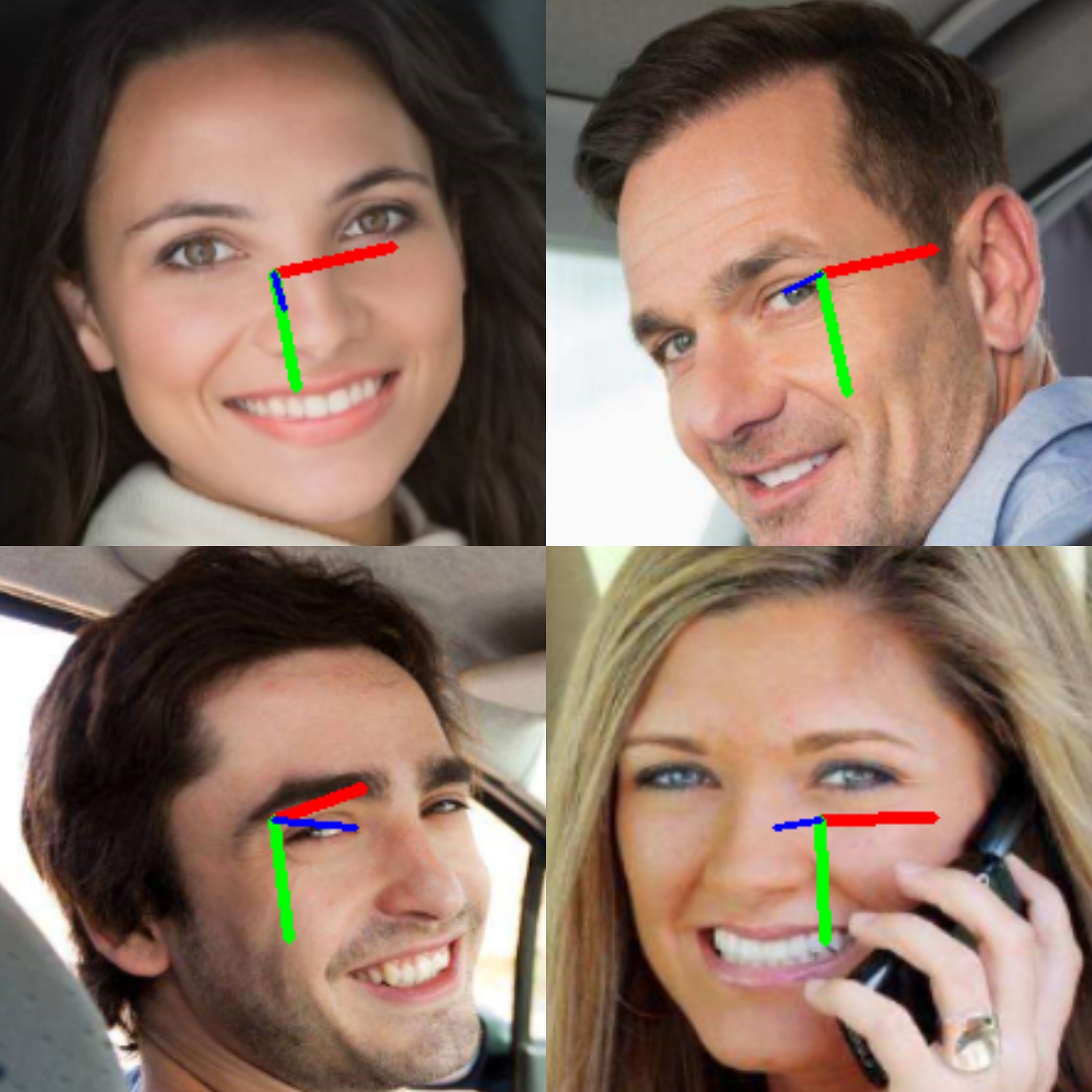}
\caption{Example pose estimation using our method. The blue axis points towards the front of the face, green pointing downward and red pointing to the side.}
\end {figure}

Recently, a fined-grained head pose estimation method\cite{ruiz2017fine} without landmarks has been proposed. It predict head pose Euler angle directly from image using a multi-loss network, where three angles are trained together and each angle loss is of two parts: a bin classification and regression component. Classification and regression components are connected through multi-loss training. However, they do not deal with the quantization errors brought by coarse bin classification. 

In our proposed method, we pose higher restriction for bin classification, in order to get better result of regression. Based on our observation, the classification converges much faster than regression, which weakens the usefulness of multi-loss training scheme. But a direct refined bin classification may counteract the benefit of problem reduction. Therefore, we introduce a hybrid coarse-fine classification framework, which proves to not only be helpful to refined bin classification but also improve the performance of prediction. Proposed network is shown in Figure 2. The main contributions of our work are summarized as below: 

$\bullet$ Use stricter fine bin classification to reduce the error brought by coarse bin classification. 

$\bullet$ Propose our hybrid coarse-fine classification scheme to make better refined classification. 

$\bullet$ State-of-the-art performance for head pose estimation using CNN-based method on AFLW2000 and BIWI datasets, and close the gap with state-of-the-art on AFLW.

\section{Related work}
\label{sec:format}

Head pose estimation has been widely studied and diverse traditional approaches have been proposed, including Appearance Template Models \cite{huang1998face}, Detector Arrays\cite{osuna1997training} and Mainfold Embedding\cite{balasubramanian2007biased}. Until now, approaches to head pose estimation have adapted to deep neural network and been divided into two camps: landmark-based and landmark-free.

Landmark-based methods utilize facial landmarks to fit a standard 3D face. 3DDFA \cite{zhu2016face} directly fits a 3D face model to RGB image via convolutional neural networks, and aligns facial landmarks using a dense 3D model, 3D head pose is produced in the 3D fitting process. SolvePnP tool \cite{gao2003complete} also produces head pose in analogous way. However, this method usually use a mean 3D human face model which introduces intrinsic error during the fitting process.

Another recent work done by Aryaman et al. \cite{gupta2018nose} achieves great performance on public datasets. They propose to use a higher level representation to regress the head pose while using deep learning architectures. They use the uncertainty maps in the form of 2D soft localization heatmap images over selected 5 facial landmarks, and pass them through an convolutional neural network as input channels to regress the head pose. However, this approach still cannot avoid problem of landmark invisibility even though they use coarse location, especially when considering that their method only involves with five landmarks which make their method very fragile to invisible condition.

Landmark-free methods treat head pose estimation as a sub-problem of multi-task learning process. M. Patacchiola\cite{patacchiola2017head} proposes a shallow network to estimate head pose, and provide a detailed analysis on AFLW dataset. KEPLER\cite{kumar2017kepler} uses a modified GoogleNet and adopts multi-task learning to learn facial landmarks and head pose jointly. Hyperface\cite{ranjan2016hyperface} also follows multi-task learning framework, which detect faces and gender, predict facial landmarks and head pose at once. All-In-One\cite{ranjan2017all} adds smile prediction and age estimation to the former method.

Chang et al.\cite{chang2017faceposenet} regresses 3D head pose by a simple convolutional neural network. However, they focus on face alignment and do not explicitly evaluate their method on public datasets. Ruiz et al.\cite{ruiz2017fine} is another landmark-free work which performs well recently, they divide three branches to predict each angle jointly, each branch is combined by classification and integral regression. Lathuiliere et. al\cite{lathuiliere2017deep} proposed a CNN-based model with a Gaussian mixture of linear inverse regressions to regress head pose. Drouard et. al\cite{drouard2017robust} further exploits to deal with issues in \cite{lathuiliere2017deep}, including illumination, variability in face orientation and in appearance, etc. by combining the qualities of unsupervised manifold learning and inverse regressions. 

Although recent state-of-the-art landmark-based method has better prediction given the ground truth, they suffer from landmark invisibility and landmark inaccuracy under real scene. Robust landmark-free method introduces extra error which limits its performance. In our work, we follow the landmark-free scheme and propose hybrid coarse-fine classification scheme which intends to solve the problem of extra error introduced by coarse classification in \cite{ruiz2017fine}.

\begin{figure*}[h]
\centering
\includegraphics[scale=0.52]{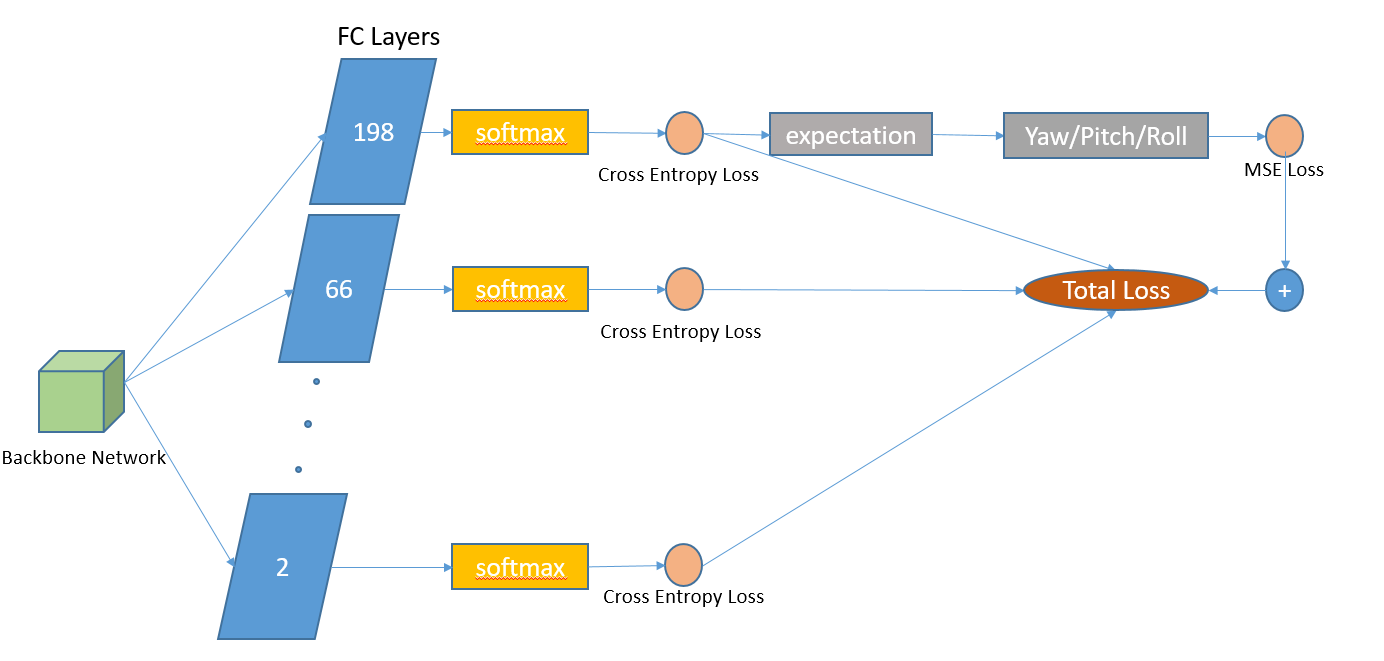}
\caption{Hybrid Coarse-fine Classification for head pose estimation. The green cude is backbone network used to extract feature, the number e.g. 198, 66, 2 in the parallelogram (FC layer) is the number of node. The upper classification branch is fine classification and used for get final prediction through integrate regression. Other branches are coarse classification and used to assist the learning. The total loss is combined by several classification loss and one regression loss as shown in equation \eqref{eq.1}.}
\end{figure*}

\section{proposed method}
\label{sec:majhead}

\subsection{Hybrid coarse-fine classification}
\label{ssec:subhead}
Although \cite{ruiz2017fine} contribute great work on head pose estimation using landmark-free method, but their work still meet with some issues. They do coarse bin classification before integrate regression. Bin classification relaxes a strict regression problem into a coarse classification problem, meanwhile, it introduces extra error which limits the performance of precise prediction. 

Multi-task learning (MTL) has led to success in many applications of machine learning. \cite{ruder2017overview} demonstrates that MTL can be treated as implicit data augmentation, representation bias, regularization and etc. Hard parameter sharing which is common in MTL, shares low representation but has task specific layer for high representation. Most of existed multi-task methods combine several related but different tasks together, such as age, gender and emotion, to learn a more general high level representation. However, hybrid classification scheme for single same task with different dimension still does not receive enough attention as far as we know.

Here, we introduce our general hybrid coarse-fine classification scheme into network, architecture is shown in Figure 2. Hybrid scheme can be regarded as a new type of hard parameters sharing, but unlike former methods which are of different tasks, each classification branch is a related and same task with specific restriction scale. It shares advantages of MTL. First, it is helpful to reduce the risk of overfitting, as the more tasks we are learning simultaneously, the more our model has to find a universal representation that captures all of the tasks and the less is our chance of overfitting on a single fine classification task. Besides, coarse classification poses with less restriction can converge faster, thus, it can help avoid some flagrant mistakes, e.g. predict a wrong sign symbol, and make the prediction more stable.
 
We use more refined classification at the highest level, which can improve the regression accuracy to bits in theory, but this operation may counteract the benefit of problem reduction. Thus, we propose our hybrid coarse-fine classification scheme to offset the influence of refined classification. The problem is relaxed multiple times on different scales in order to ensure precise prediction under different classification scale. We take both coarse bin classification and relatively fine bin classification into account, each FC layer represents a different classification scale and compute its own cross-entropy loss. In the integrate regression component, we only utilize result of the most refined bin classification to compute expectation and regression loss. One regression loss and multiple classification losses are combined as a total loss. Each angle has such a combined loss and share the previous convolutional layers of the network.

Our proposed hybrid coarse-fine classification scheme can be easily added into former framework and bring performance up without much extra computing resources. The final loss for each angle is the following:

\begin{equation}
\label{eq.1}
Loss = \alpha * MSE(y,y^{*}) + \sum_{i=1}^{num} {\beta_{i}*H(y_{i},y^{*}_{i})}
\end{equation}

where H and MSE respectively designate the cross-entropy loss and mean squared error loss functions, num means the number of classification branch which is set to 5 in our case.

We have tested different coefficients for the regression component and hybrid classification component, our results are presented in Table 4. and Table 5.

\subsection{Integrate regression}
\label{ssec:subhead}

Xiao et al.\cite{sun2018integral} introduces integrate regression into human pose estimation to cope with non-differentiable post-processing and quantization error. Their work shows that a simple integral operation relates and unifies the heat map representation and joint regression. Ruiz et al.\cite{ruiz2017fine} utilizes integrate regression in head pose estimation. 

This scheme treats a direct regression problem as two steps process, a multi-class classification followed by integrate regression, by modifying the “taking-maximum” of classification to “taking-expectation”, and a fine-grained predictions is obtained by computing the expectation of each output probability for the binned output.

We follow this setting in our network and use the same backbone network as \cite{ruiz2017fine} in order to fairly compare. Intuitively, such scheme can be seen as a way of problem reduction, as bin classification is a coarse annotation rather than precise label, classification and the output are connected through multi-loss learning which makes the classification also sensitive the output. Another explanation is that bin classification uses the very stable softmax layer and cross-entropy loss, thus the network learns to predict the neighborhood of the pose in a robust fashion.

\section{Experiments}
\label{sec:majhead}
\subsection{Datasets for Pose Estimation}
\label{ssec:subhead}
We demostrate that datasets under real scene with precise head pose annotations, numerous variation on pose scale and lighting condition, is essential to make progress in this filed. Three benchmarks are used in our experiments.

300W-LP \cite{zhu2016face}: is a synthetically expanded dataset, and a collection of popular in-the-wild 2D landmark datasets which have been re-annotated. It contains 61,225 samples across large poses, which is further expanded to 122,450 samples with flipping.

AFLW2000 \cite{koestinger2011annotated}: contains the first 2000 identities of the in-the-wild AFLW dataset, all of them have been re-annotated with 68 3D landmarks.

AFLW \cite{koestinger2011annotated}: contains 21,080 in-the-wild faces with large-pose variations (yaw from -90$^\circ$ to 90$^\circ$).

BIWI \cite{fanelli2011real}: is captured in well controlled laboratory environment by record RGB-D video of different people across different head pose range using a Kinect v2 device and has better pose annotations. It contains about 15, 000 images with $\pm$75$^\circ$ for yaw, $\pm$60$^\circ$ for pitch and $\pm$50$^\circ$ for roll. 

\begin{table}[htbp]
\centering
\begin{tabular}{lllll}
\hline \hline
Method            & Yaw    & Pitch  & Roll   & MAE    \\
\hline
3DDFA{}\cite{zhu2016face}{}     & 5.400  & 8.530  & 8.250  & 7.393  \\
\hline 
Ruiz etal.{}\cite{ruiz2017fine}{} & 6.470  & 6.559  & 5.436  & 6.155  \\
\hline 
\textbf{Ours}              & \textbf{4.820}  & \textbf{6.227}  & \textbf{5.137}  &
\textbf{5.395}\\
\hline  \hline 
\end{tabular}
\caption{Mean average error of Euler angles across different methods on the AFLW2000 dataset.}
\end{table}

\subsection{Pose Estimation on the AFLW2000}
\label{sssec:subsubhead}
Same backbone network as \cite{ruiz2017fine} is adopted. Network was trained for 25 epochs on 300L-WP using Adam optimization \cite{kinga2015method} with a learning rate of 10$^{-6}$ and $\beta_{1}$ = 0.9, $\beta_{2}$ = 0.999 and ε = 10$^{-8}$. We normalize the data before training by using the ImageNet mean and standard deviation for each color channel. Our method bins angles in the $\pm$99$^\circ$ range we discard images with angles outside of this range. Results can be seen in Table 1.

\begin{table}[htbp]
\centering
\begin{tabular}{lllll}
\hline \hline 
Method                & Yaw    & Pitch  & Roll   & MAE    \\
\hline 
Liu et al.{}\cite{liu20163d}{}    & 6.0    & 6.1    & 5.7    & 5.94   \\
\hline 
Ruiz et al.{}\cite{ruiz2017fine}{}    & 4.810  & 6.606  & 3.269  & 4.895  \\
\hline 
Drounard{}\cite{drouard2017robust}{}      & 4.24   & 5.43   & 4.13   & 4.60   \\
\hline 
DMLIR{}\cite{lathuiliere2017deep}{}         & \textbf{3.12}   & 4.68   & 3.07   & 3.62   \\
\hline 
MLP + Location{}\cite{gupta2018nose}{} & 3.64   & 4.42   & 3.19   & 3.75   \\
\hline 
CNN + Heatmap{}\cite{gupta2018nose}{}  & 3.46   & 3.49   & \textbf{2.74}   & 3.23   \\
\hline 
\textbf{Ours}                  & 3.4273 & \textbf{2.6437} & 2.9811 & \textbf{3.0174}\\
\hline \hline
\end{tabular}
\caption{Mean average error of Euler angles across different methods on the BIWI dataset with 8-fold cross-validation.}
\end{table}

\begin{table}[htbp]
\centering
\begin{tabular}{lllll}
\hline \hline 
Method                    & Yaw   & Pitch & Roll & MAE   \\
\hline 
Patacchiola et al.{}\cite{patacchiola2017head}{} & 11.04 & 7.15  & 4.40 & 7.530 \\
\hline 
KEPLER{}\cite{kumar2017kepler}{}             & 6.45  & 5.85  & 8.75 & 7.017 \\
\hline 
Ruiz et al.{}\cite{ruiz2017fine}{}        & 6.26  & 5.89  & 3.82 & 5.324 \\
\hline 
MLP + Location{}\cite{gupta2018nose}{}     & 6.02  & 5.84  & 3.56 & 5.14  \\
\hline 
\textbf{Ours}                      & 6.18  & 5.38  & 3.71 & 5.090  \\
\hline 
CNN + Heatmap{}\cite{gupta2018nose}{}      & 5.22  & 4.43  & 2.53 & 4.06 \\
\hline \hline
\end{tabular}
\caption{Mean average error of Euler angles across different methods on the AFLW dataset.}
\end{table}

\subsection{Pose Estimation on the AFLW and BIWI Datasets}
\label{sssec:subsubhead}
We also test our method on AFLW and BIWI dataset with same parameter setting as 4.2. Results can be seen in Table 2. and Table 3. Our method achieves state-of-the-art on BIWI, MAE of our method is lower the base network \cite{ruiz2017fine} by large scale, and also achieves better performance than the recent landmark-based CNN + Heatmap \cite{gupta2018nose} method. We demonstrate that our method perform well on BIWI as it is captured in controlled environment, and has better ground truth annotations, but this verifies the usefulness of our hybrid  coarse-fine classification scheme when the annotation is precise.

We also surpass all landmark-free methods, and achieve competing performance over all methods on AFLW, following testing protocol in \cite{kumar2017kepler} (i.e. selecting 1000 images from testing and remaining for training.)

\subsection{AFLW2000 Multi-Classification Ablation}
\label{sssec:subsubhead}

\begin{table}[htbp]
\centering 
\begin{tabular}{llllllllll}
\hline  \hline 
$\alpha$  & $\beta_{1}$ & $\beta_{2}$ & $\beta_{3}$ & $\beta_{4}$ & $\beta_{5}$ & MAE    \\
\hline 
2 & 1 & 0 & 0 & 0 & 0 & 5.7062 \\
\hline 
2 & 3 & 1 & 1 & 1 & 1  & 5.6270 \\
\hline 
2 & 5 & 3 & 1 & 1 & 1 & 5.6898 \\
\hline 
2 & 7 & 5 & 3 & 1 & 1  & \textbf{5.3953} \\
\hline 
2 & 9 & 7 & 5 & 3 & 1 & 5.5149 \\
\hline \hline
\end{tabular}
\caption{Ablation analysis: MAE across different classification loss weights on the AFLW2000 dataset.}
\end{table}

\begin{table}[htbp]
\centering 
\begin{tabular}{llllllllll}
\hline \hline
$\alpha$   & $\beta_{1}$ & $\beta_{2}$ & $\beta_{3}$ & $\beta_{4}$ & $\beta_{5}$ &  MAE    \\
\hline
0.1 & 7 & 5 & 3 & 1 & 1 &  5.4834 \\
\hline
1   & 7 & 5 & 3 & 1 & 1 &  5.6160 \\
\hline
2   & 7 & 5 & 3 & 1 & 1 &  \textbf{5.3953} \\
\hline
4   & 7 & 5 & 3 & 1 & 1 &  5.6255 \\
\hline \hline
\end{tabular}
\caption{Ablation analysis: MAE across different regression loss weights on the AFLW2000 dataset.}
\end{table}

In this part, we present an ablation study of the hybrid coarse-fine classification. We train ResNet50 using different coefficient setting. Results can be seen in Table 4. and Table 5. We observe the best results on the AFLW2000 dataset when the coefficient is 2,7,5,3,1,1 by order. $\alpha$,$\beta_{1}$,$\beta_{2}$,$\beta_{3}$,$\beta_{4}$,$\beta_{5}$ correspond to the weight of regression, 198 classes, 66 classes, 18 classes, 6 classes and 2 classes.

\section{conclusion}
\label{sec:page}
We present a hybrid classification scheme for precise head pose estimation without facial landmarks. Our proposed method achieves state-of-the-art on BIWI and AFLW2000 dataset, and also achieves promising performance on AFLW dataset. The hybrid coarse-fine classification framework is proved to be beneficial for head pose estimating and we believe that it is not only beneficial for single specific task, it may be also helpful to other classification problem such as digits recognition, we will work on it later.

\newpage
\bibliographystyle{IEEEbib}
\bibliography{r1,r2,r3,r4,r5,r6,r7,r8,r9,r10,r11,r12,r13,r14,r15,r16,r17,r18,r19,r20,r21,r22,r23,r24,r25,r26}
\end{CJK*}
\end{document}